\documentclass[runningheads]{llncs}
\usepackage{amsfonts}

\usepackage{graphicx}
\begin{document}
\title{An information theoretic approach to the autoencoder}
%
%
\author{Vincenzo Crescimanna \and
Bruce Graham}
\authorrunning{Crescimanna V. and Graham B.}
%
\institute{University of Stirling, Stirling, UK\\
\email{vincenzo.crescimanna1@stir.ac.uk}\\
}
\maketitle              
\begin{abstract}
We present a variation of the Autoencoder (AE) that explicitly maximizes the mutual information between the input data and the hidden representation. The proposed model, the InfoMax Autoencoder (IMAE), by construction is able to learn a robust representation and good prototypes of the data. IMAE is compared both theoretically and then computationally with the state of the art models: the Denoising and Contractive Autoencoders in the one-hidden layer setting and the Variational Autoencoder in the multi-layer case. Computational experiments are performed with the MNIST and Fashion-MNIST datasets and demonstrate particularly the strong clusterization performance of IMAE.

\end{abstract}
\section{Introduction}
Nowadays, Deep Neural Networks (DNNs) are considered the best machine learning structure for many different tasks. The principal property of this family is the ability to learn a representation of the input data in the hidden layers. 
Indeed, as underlined in \cite{ben_replen}, in order to understand the world, an intelligent system needs to learn a compact representation of the high dimensional input data. 

In biological Neural Networks (NNs) processing sensory information, it is hypothesised that the first layers encode a task-independent representation that maximizes the mutual information with the input \cite{barlow}.
Given that it is desirable to learn a representation that is able to identify and disentangle the underlying explanatory hidden factors in input data, and inspired by the behaviour of real NNs, we propose an unsupervised Infomax Autoencoder (IMAE) NN that maximizes the mutual information between the encoded representation and the input. The results of computational experiments show that IMAE is both robust to noise and able to identify the principal features of the data.

The paper is structured as follows. In the next section we describe related work. Section 3 describes the derivation of the model, which is based on a classic approximation of differential entropy. Due to comparison with other AEs, we give also a geometric and a generative description of the model. The theoretical differences are confirmed in the fourth section by computational experiments, where  the clusterization (representation) and the robustness of IMAE are evaluated.  The work is concluded with a brief comment on the results and a description of future work.

\section{Related Work}
The first neural network defined to explicitly maximize the information between the input and the hidden layer was proposed by Linsker \cite{linsker}, giving to the objective function the name, \emph{Infomax principle}. This model is a linear model that actually maximizes only the entropy of the representation, performing Principal Component Analysis (PCA) in the case of Gaussian distributed data.  The same objective function was applied by Bell and Sejnowsky \cite{bell95info}, showing that a 1-layer neural network with sigmoid activation performs  Independent Component Analysis (ICA).

Both these models are quite restrictive, indeed they work only under the assumption that the visible data is a linear combination of hidden features. 
A more general way to extract hidden features is given by the Autoencoder (AE), a NN that is a composition of an encoder and a decoder map, respectively $W$ and $V$. This model can be seen as a generalization of the PCA model, because in the assumption $W$ and $V$ are linear maps, the space spanned by $W$ is the same of the one spanned by principal components, see. e.g. \cite{Baldi}.

An information theoretic description of an AE was given by Vincent et al. \cite{Vincent1} where, with restrictive assumptions, they observed that reducing the reconstruction loss of an AE  is related to maximizing the mutual information between the visible and hidden variables. 

The IMAE, as we will see in the next section, is an AE that is able to learn a robust representation. In the literature there are many models in this family that are defined to learn a good representation, see e.g. \cite{karhunen2015} and references therein. For practical reasons, in the following section we compare the IMAE with the AEs that are recognized to learn the best features: Denoising, Contractive and Variational Autoencoders.

\section{Infomax autoencoder model}
Shannon \cite{Shannon} developed Information theory in order to define a measure of the efficiency and reliability of signal.  In this theory a key role is assumed by Mutual Information $I$, i.e. a measure of the information shared between two variables, in the signal transmission case: the original message and the received one.

Formally, given two random variables: $X$ and $Y$, the mutual information between these two variables is defined as \begin{equation}\label{eq_IM}
    I(X,Y) = h(Y) - h(Y|X)
\end{equation}
where $h(Y)$ is the (differential) entropy of $Y$, a measure of the average amount of information conveyed, and $h(Y|X)$ is the conditional entropy $Y$ with respect to $X$, a measure of the average of the information conveyed by $Y$ given $X$, i.e. the information about the processing noise, rather than about the system input $X$. 

Given the assumption the hidden representation $Y$ lives in a manifold embedded in a subspace of the visible space, by the definition above, the Infomax objective function finds the projection from the visible (higher dimension space) to the hidden manifold that preserves as much information as possible. Defining $W$ as the projection map, the Infomax objective can be state as: $\max_W I(X, W(X))$, with the representation, $Y = W(X)$, that maximizes the mutual information with the system input $X$. 

In general, to compute the mutual information between two variables is not trivial, indeed it is necessary to know the distribution of the input and of the hidden representation, and to compute an integral. For this reason, we propose an approximation of the mutual information $I(X,Y)$.

In the following, capital letters $X, \hat{X}, Y$ denote the random variables and lowercase letters $x, \hat{x} \in \mathbb{R}^d, y \in \mathbb{R}^l $ the respective realisations. The map between the input space $\mathbb{R}^d$ to the hidden representation $\mathbb{R}^l$ is $W, Y = W(X) $, and the map from $\mathbb{R}^l$ to the input space is $V, \hat{X} = V(Y)$.
In the first step we assume that the map $W$ is invertible, then both $x, y \in \mathbb{R}^d$. 

Our approximation starts by considering the conditional entropy in eq (\ref{eq_IM}).  In order to estimate this quantity it is necessary to suppose there exits an unbiased efficient estimator $\hat{X} = V(Y) $ of the input $X$, in this way the conditional entropy $h(Y|X)$ can be approximated by $h(\hat{X}|X)$.
Indeed, the conditional entropy is the information about the processing noise, and the noise added in the encoding process is proportional to the noise between the input and its reconstruction. Formally,  see e.g. \cite{papa} p. 565, in the case of a bijiective function the following holds  
\begin{equation} \label{_ineq}
h(\hat{X}|X) \leq h(Y|X) + \int \log |J_V(y)| dp(y), 
\end{equation}
where the equality holds in the case that $V$ is an orthogonal transformation, with $J_V$ the Jacobian of the function $V$ and $|\cdot| = |\det(\cdot)| $ .

Let us start by assuming the second term on the right-hand-side (RHS) is bounded (in the end we show that this term is bounded). 
In order to have the inequality as tight as possible, the idea is to assume the conditional variable $\hat{X}|X$ is Gaussian distributed with mean $X$ and variance $E[(X-\hat{X}) (X-\hat{X})^T]$. Indeed, the Gaussian distribution has maximum entropy.
Since $\hat{X}|X$ is Gaussian distributed, the conditional entropy $h(\hat{X}|X)$ is an (increasing) function of the covariance matrix $\Sigma_{\hat{x}|x}$, i.e. $h(\hat{X}|X) \propto \mbox{const} +\log(|\Sigma_{\hat{x}|x}|)$. 
By the symmetric definition of the covariance matrix and the well known relationship $log(|\Sigma|) = trace(\log(\Sigma))$, it is sufficient to consider as the loss function to minimize the trace of $\Sigma_{\hat{x}|x}$, i.e. the $L_2$-norm of the input with its reconstruction. In this way, we obtain formally that $V$ should be near, in the $L_2$ space of functions, to the inverse of $W$, i.e. $V\circ W \approx Id$; then also the Jacobian $J_{V \circ W} \approx Id$.

Now let us focus on the first term of the RHS of (\ref{eq_IM}), the differential entropy $h(Y)$. Defining the function $W$ as a composition of two functions: $W = \sigma \circ W_0$, the entropy function can be written as follows: 
\begin{equation}
    h(Y) = \int\log |J_\sigma(Y_0)|p(y_0) + h(Y_0)
\end{equation}
where $Y_0 = W_0(X)$. In our application, we consider the non-linearity $\sigma$ as the logistic function $\sigma(x) = (1+\exp(-x))^{-1}$; this allows to have the representation $Y_0$ distributed following the logistic distribution, see \cite{bell95info}. 
In order to compute the second term, $h(Y_0)$,  following \cite{hyvrinen1998new,hyvarinen2000independent}, this term can be seen as a sparse penalty in the $Y_0$ output, that, as observed in \cite{lee_ica}, encourages to have an independent distribution of the components of $Y_0$.
Since the $Y_0$ components are almost independent, a good approximation for the determinant of the Jacobian of the sigmoid is the product of its diagonal elements. Then observing that the variance of $Y_0$ is controlled by the Jacobian of the sigmoid, the entropy of $Y$, can be approximated as: \begin{equation}\label{ent}
    h(Y) = \sum_i \sigma_i(Y_0)(1-\sigma_i(Y_0)) - (\log(\cosh(Y_0)_i))^2,
\end{equation}
where the first term in the RHS is associated to the $\log$ of the determinant sigmoid Jacobian, and the second term is the sparsity penalty, associated to $h(Y_0)$.
The sparsity penalty in the equation above ensures $W$ to be a bounded operator and, by the open mapping theorem, $J_W$ and $J_V$ have bounded determinants different from zero, thus the inequality (\ref{_ineq}) is well defined.

\paragraph{Generalization to under- and over-complete representations.}

The approximation of the conditional entropy as the $L_2$ reconstruction loss was derived under the condition that the input and the representation live in the same space.
A way to generalize the derivation to the under- and over-complete setting is thinking of this structure as a noisy channel, where some units (components) in the hidden and input layers are masked. Indeed, since by definition $V$ behaves as the inverse of the (corrupted) map $W$, it is not adding more, or even different, noise than that already added in the map $W$.  Thus, we can conclude that the approximation $h(Y|X) \approx h(\hat{X}|X)$ holds both in the over- and under- complete setting. By construction, the approximation of the entropy term does not have a particular requirement, so can be described for a generic number of hidden units.

\paragraph{Properties.}
The difference of IMAE from the other models in the literature is given by the latent loss term i.e. the maximization of the latent entropy. Choosing the sigmoid activation in the latent layer, in accordance with the theory of Independent Component Analysis and processing signals \cite{bookhyva}, we suppose that the latent distribution is logistic and concentrated around the mean. In particular, the first term of the entropy (\ref{ent}), ensures that the distribution is peaked around the mean value, guaranteeing the learning of good prototypes of the input data. Instead, the sparseness penalty encourages the independence between the variables, guaranteeing a model robust to noise.

\subsubsection{Relationship with other approaches.} In order to describe the properties of IMAE from different perspectives, we compare it with the most common AE variants able to learn good representations. In all the cases that we consider the encoder map has the form $\sigma \circ W_0$.

\paragraph{Contractive Autoencoder (CAE).}
Starting from the idea that a low-derivative hidden representation is more robust, because it is less susceptible to small noise, Rifai et al. \cite{rifai} suggested a model that explicitly reduces (contracts) the Jacobian of the  encoder function.  The proposed Contractive Autoencoder (CAE) has a structure similar to IMAE, with sigmoid activation in the hidden layer, but the latent loss is the Froebenius norm of the encoder Jacobian $J_W$.
In the case $W_0 $ is a linear matrix, the latent loss can be written as 
\begin{equation}\label{rif_eq}
	\| J_W(x) \|_F^2 = \sum_i^l (\sigma_i(1-\sigma_i))^2\sum_j^d (W_0)_{ij}^2.
\end{equation}
From eq (\ref{rif_eq}), it is clear that CAE encourages a flat hidden distribution, the opposite of IMAE, that encourages a big derivative around the mean value; the consequences of such differences will be underlined and clarified in the computational experiment section, below. 
From an information theory perspective, observing that the Froebenius norm is an approximation of the absolute value of the determinant, the CAE representation can be described as a low entropy one.  Indeed, by changing variables in the formula, in the case of a complete representation, the entropy of $Y$ is a linear function of the log-determinant of the Jacobian of $W$.

\paragraph{Denoising Autoencoder (DAE).}
Assuming that robust reconstruction of the input data implies robust representation, Vincent et al. \cite{Vincent1} proposed a variant of the AE trained with corrupted input, termed the Denoising Autoencoder (DAE). From a manifold learning perspective, Vincent et al. observed that the DAE is able to learn the latent manifold, because it learns the projection from the corrupted  $\tilde{X}$ to the original $X$ input. The problem, as highlighted in section 4, is that there is no assumption on the distribution of the data around the hidden manifold i.e. DAE learns only a particular type of projection and then is robust to only one type of noise. 
Differently in IMAE, the preferred projection direction is suggested by the choice of the hidden activation. 

\paragraph{Variational Autoencoder (VAE).} The Variational Autoencoder (VAE) \cite{kigma} is not a proper neural network, but actually a probabilistic model; differently from the other models described until now it does not discover a representation of the hidden layer, but rather the parameters $\lambda$ that characterize a certain distribution chosen a-priori $q_\lambda(y_0|x)$ as a model of the unknown probability $p(y_0|x)$. In the case that $q_\lambda$ is Gaussian, the parameters are only the mean and variance and VAE can be written as a classic AE with loss function: 
\begin{eqnarray}
	\mathcal{L}_{VAE} = - E_q(\log p(x|y_0)) + D_{KL}(q_\lambda(y_0|x)\| p(y_0)).
\end{eqnarray}
The first term of $\mathcal{L}_{VAE}$ is approximately the reconstruction loss. The Kullback-Leibler (K-L) divergence term, $D_{KL}$, which is the divergence in the probability space between the a-priori and the real distributions, represents the latent loss of the neural network, approximated in our case as: $\sum_i \mu(y_0)_i^2 + \sigma(y_0)_i^2 - \log(\sigma(y_0)_i^2) -1. $

From the IMAE perspective, we observe that the first term of the latent loss, $-h(Y)$,  encourages the $Y_0$-distribution to be logistic with small variance, and the second term penalizes the dependencies between the $Y_0$ components.
Then we can conclude that, from a probabilistic perspective, the IMAE latent loss works like a K-L divergence, with the target function a product of independent logistic distributions. 

\section{Computational experiments}
In this section we describe the robustness and the clusterization performance of IMAE, comparing it with the models described above and the classic AE. We carry out the tests with two different architectures: single- and multi-hidden layers.  The first comparison is made with AE, CAE and DAE; in the second comparison, which is suitable for a small representation, we compare IMAE with VAE.

\subsection{1-hidden layer}

The first tests are carried using the MNIST dataset, a collection of $28\times 28$ grey-scale handwritten digits, subdivided into ten classes, with 60 000 training and 10 000 test images.
We consider two different AutoEncoder architectures, one under-complete with 200  hidden units and the other over-complete with 1000 hidden units.  All the neural networks are trained using batch Gradient Descent with learning rate equal to 0.05 and batch size composed by 500 input patterns (images), trained for 2000 epochs. For all the models, the encoder and decoder functions have form: $W(x) = \sigma(W_0x)$ and $V(x) = Vx$. The listed results are obtained with tied weights, $W_0 = V^T$, but equivalent results are obtained with $W$ and $V$ not related to each other. 
The DAE is trained both with Gaussian (DAE-g) and mask (DAE-b) noise. In DAE-g, each pixel of the training images is corrupted with zero-mean Gaussian noise with standard deviation $\sigma = 0.3$; in DAE-b, each pixel is set to zero with probability $p = 0.3$ . In CAE and IMAE, the hyper-parameters associated with the latent loss are set equal to $0.1$ and $1$, respectively.

\paragraph{Robust reconstruction.}
As suggested by \cite{Vincent1}, a way to evaluate the robustness of the representation is by evaluating the robustness of the reconstruction.
Under this assumption, for each model we compute the $L_2$-loss between the reconstructed image and the original (clean) one. Results for this measure are listed in Table \ref{l2_MNIST}.
From these results we can deduce that IMAE is equally robust in all the considered settings and, excluding the respective DAEs trained with the test noise, has the best reconstruction performance. In contrast, DAE is robust only to the noise with respect to which it was trained and CAE is only robust in the over-complete setting with respect to Gaussian noise. The classic AE is best in the zero-noise case.

\begin{table}[h!]
    \parbox{.45\linewidth}{
    \centering
    \begin{tabular}{c|c c c c}
     \multicolumn{5}{c}{Mask noise, $p$}\\
    \hline 
        Model & 0 & 0.3 & 0.5 & 0.75  \\
        \hline\hline
         AE & \textbf{12.8} & 110.2 & 145.7 & 183.1\\
         CAE & 129.4 & {140} & 157.7 & 254.3\\
         DAE-b & 70.5 & \textbf{68.9} & \textbf{97.6} & \textbf{151.7}\\
         DAE-g & 263.3 & {247.4} & {283.6} & {419.6}\\
         IMAE & 53.7 & 106.5 &132.9 &{161.8}\\
    \hline
    \end{tabular}
    }
    \hfill
    \parbox{.45\linewidth}{
    \centering
    \begin{tabular}{c|c c c c}
    \multicolumn{5}{c}{Gaussian noise, $\sigma$}\\
    \hline 
        Model & 0.03 & 0.15 & 0.35 & 0.45  \\
        \hline\hline
         AE & \textbf{44.3} & {130.5} & 178.1 & 190.9\\
         CAE & 133.2 & {149.1} & 156.9 & 159.5\\
         DAE-b & 86.6 & {165.4} & {210} & 226.1\\
         DAE-g & 202.4 & \textbf{78.6} & \textbf{77.5} & \textbf{92}\\
         IMAE & 73.4 & {126.9} & {156.6} & {166.4}\\
    \hline
    \end{tabular}
    }
    \vspace{.75cm}\\
    
    \parbox{.45\linewidth}{
    \centering
    \begin{tabular}{c|c c c c}
    \hline 
        Model & 0 & 0.3 & 0.5 & 0.75  \\
        \hline\hline
         AE & \textbf{37.4} & 97.4 & 133. & 176.3\\
         CAE & 141.1 & {148.3} & 157.7 & 206.4\\
         DAE-b & 71.3 & \textbf{70.5} & \textbf{97.1} & \textbf{152}\\
         DAE-g & 158.2 & {148.7} & {167.2} & {226.4}\\
         IMAE & 103.8 & 125 &138.9 &{155.8}\\
    \hline
    \end{tabular}
    }
    \hfill
    \parbox{.45\linewidth}{
    \centering
    \begin{tabular}{c|c c c c}
    \hline 
        Model & 0.03 & 0.15 & 0.35 & 0.45  \\
        \hline\hline
         AE & \textbf{50.4} & {122.2} & 163.6 & 176.6\\
         CAE & 143.4 & {151.6} & {158.5} & 160.7\\
         DAE-b & 86.6 & {154.6} & {193} & 205.3\\
         DAE-g & 133.5 & \textbf{72.2} & \textbf{77.9} & \textbf{91.5}\\
         IMAE & 112.5 & {135.7} & {148.1} &{152.7}\\
    \hline
    \end{tabular}
    }
     \vspace{.25cm}
    \caption{Reconstruction loss performance with MNIST input data corrupted with masking noise with probability $p = 0, 0.3, 0.5, 0.75$ (left column) or with zero-mean Gaussian noise with standard deviation $\sigma = 0.03, 0.015, 0.35, 0.45$ (right column), 1000 hidden units (above), 200 hidden units (below).}
    \label{l2_MNIST}
\end{table}

\paragraph{Prototype identification.}
A good, task-independent representation should include the hidden prototypes for each class. Assuming that the hidden representation of each element lies close to the respective prototype, to evaluate representation quality we measure the clusterization performance.  In particular, we compute the Rand-index between the ten clusters identified by the K-means algorithm on the hidden space and the ones defined by ground-truth in the MNIST data set. Formally we compute the Rand-index $R$ as follows: 
\begin{equation}
    R = \max_m \frac{\sum_{i=1}^N \mathbf{1}(m(c_i) = l_i)}{N}
\end{equation}  
where $c_i\in \{1, \dots, k\}$ is the hidden representation cluster in which the current input, $x_i$, belongs, $l_i \in \{1, \dots, k \}$ is the ground truth label associated with $x_i \in \{ x_i\}_1^N$, $m$ is a 1-1 map from cluster to label,  and $\mathbf{1}$ is the indicator function, that is 1 if $m(c_i) = l_i$, 0 otherwise. In our case $k = 10$, the number of different classes, and $N = 1000$; the results are shown in Table \ref{clust}.
   
\begin{table}[]
    \centering
     \parbox{.45\linewidth}{
    \centering
        \begin{tabular}{p{.5cm}||p{.95cm}|p{.95cm}|p{1cm}| p{1cm}| p{.95cm}|}
         &  AE & CAE & DAE-b & DAE-g & IMAE\\
         \hline
      $R$  &  53.5 & 17.9 & {53.8} & 51.3 & \textbf{54.4}\\
       \hline
        $R_\nu$  &  53.5 & 17.8& 54 & 50.2 & \textbf{55.2}\\
       \hline
$\sigma'$ & 1.9  & - & 1.5 & 1.11 & 5.9  \\
\hline
    \end{tabular}
        200 hidden units
    }
    \hfill
    \parbox{.45\linewidth}{
    \centering
    \begin{tabular}{p{.5cm}||p{.95cm}|p{.95cm}|p{1cm}| p{1cm}| p{.95cm}|}
         &  AE & CAE & DAE-b & DAE-g & IMAE\\
         \hline
      $R$  &  55.7 & 17.3 & 55.6 & 51.4 & \textbf{55.8}\\
       \hline
        $R_\nu$  &  55.6 & 17.3 & {55.3} & 50.5 & \textbf{55.6}\\
       \hline
$\sigma'$ & 3.9 & -  & 4.5 & 2.5 & 5.8  \\
\hline
    \end{tabular}\\
        {1000 hidden units}

    }
    \vspace{.3cm}
    \caption{Averave over 50 iterations of the Rand indices obtained with clean, $R$, and noisy, $R_\nu$  ($\nu \sim \mathcal{N}(0, (0.2)^2)$), test data. The $\sigma'$ row denotes the average derivative of the hidden representation (the order of the derivative is $10^{-2}$; note all the CAE derivatives are around the order $10^{-5}$). } 
    \label{clust}
\end{table}

IMAE has the best clusterization performance (high Rand index) and is also the more robust to noise. In particular, we observe that these performances are associated with the mean derivative of the non-linear hidden activation, where  
a high derivative is associated with good performance. 
This observation leads us to conclude that the derivative value can be considered as an empirical measure of the clusterization performance. 

Concerning both sets of results, we underline that reconstruction robustness is not associated with robust clusterization. For example, according to the dropout idea, DAE-b has a more robust representation than DAE-g also in the case of the Gaussian noise corruption on which it was not trained.

\subsection{Multi-hidden layers}
In many applications a shallow network structure is not sufficient, because the considered data are more redundant than the MNIST data set and the desired representation is really small. For this reason it is important to consider a deeper structure. In this setting AE, CAE and DAE cannot naturally be extended, so we compare IMAE with VAE, the AE that, better than others, is able to learn a hidden representation. 
We now consider a network with hidden architecture of type: 1100-700-$n_h$-700-1100, with softplus activation in each hidden layer except for the $n_h$ units, where a sigmoid nonlinearity is applied. In the following, we use hidden units to refer to the $n_h$-variables, where $n_h \in \{5, 10, 20\}$.
The tests are carried out with MNIST and Fashion-MNIST (F-MNIST), a MNIST-like dataset where the digits are replaced by images of clothes, \cite{fmnist}. This dataset is suitable for this network structure since it is more redundant than MNIST but still low-dimensional. The network is trained as in the 1-hidden layer architecture, with learning rate equal to $5\cdot 10^{-3}$.

Since we are interested to learn a small representation of the data, we compare the clusterization performance of VAE and IMAE. We do not consider the robustness of reconstruction because, by construction, the models are robust to really small noise.

\paragraph{Clusterization Performance.} Table \ref{l2_MNIST_tab} lists the Rand-indices obtained in the hidden space, with possibly Gaussian-noise-corrupted input data. In the noisy case the input data is corrupted with zero mean Gaussian noise with standard deviation, respectively, $\sigma = 0.01, 0.1$ for MNIST and F-MNIST. For MNIST, we choose this relatively small noise because, since it is not particularly redundant, the deep structure is not robust.

\begin{table}
	\centering
\begin{tabular}{c|c c c| c c c || c c c| c c c}
	
\multicolumn{1}{c}{}&
\multicolumn{6}{c}{MNIST} &
\multicolumn{6}{c}{F-MNIST}\\
	& & $n_h$&& &noisy&& &$n_h$&& &noisy&\\
	\hline 
	Model & 5 & 10 & 20 & 5 & 10 & 20 & 5 & 10 & 20 & 5 & 10 & 20  \\
	\hline\hline
	VAE & 70.1 & {57.2} & 52.9& 61 & {51.7} & 36 & 40.3 & 42.3 &42.4 & 38.1& 41.8& 41.3\\
	IMAE & \textbf{76.8} & \textbf{75.7} & \textbf{54.8}& \textbf{68.7} & \textbf{60.5} & \textbf{42.2}& \textbf{55.5} & \textbf{59.3} &\textbf{56} & \textbf{55}& \textbf{55.3}& \textbf{55.3}\\
	\hline
\end{tabular}
\vspace{.25cm}
\caption{Average  over 50 iterations of Rand-indices of VAE and IMAE with networks having 5, 10 or 20 hidden units. }
\label{l2_MNIST_tab}
\end{table}

\begin{figure}
    \centering
    \parbox{.45\linewidth}{
    \includegraphics[width=0.45\textwidth]{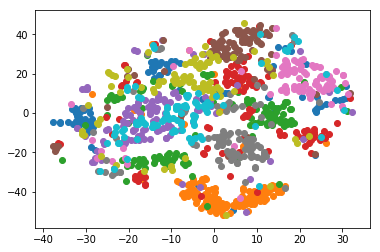}
    }
    \parbox{.45\linewidth}{
    \includegraphics[width=0.45\textwidth]{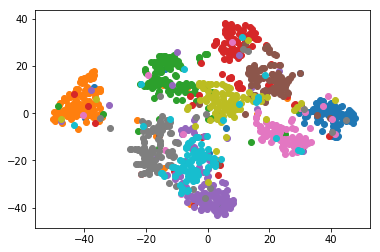}
    }
    \caption{Illustration of t-SNE encoding \cite{tsne} of VAE (left) and IMAE (right) representation obtained with noisy MNIST,  $n_h = 10$.}
    \label{tsne_hidden}
\end{figure}

The results highlight how IMAE is able to clusterize the data in a small hidden space and identify the prototypes for each class also in the case of noisy input data i.e. the hidden prototypes are robust. In order to clarify the robustness of the clusters, Figure \ref{tsne_hidden} visualises the hidden representations of the two models in the case of noisy input data. In the IMAE representation, despite the noise, it is possible to clearly distinguish the different clusters; instead in the VAE representation all the points are mixed, except largely the orange ones (bottom cluster), which represent the digit 1.

\section{Conclusions and future work}

In this paper, we derived formally an Autoencoder that explicitly maximizes the information between the input data and the hidden representation and introduces an empirical measure of the clusterization performance on the hidden space. The experiments show that IMAE is versatile, works both in shallow and deep structures, has good reconstruction performance
in the case of noisy input data,
and is able to learn good and robust prototypes in the hidden space. 
The latter point, that is also the principal innovation, is particularly useful for applications, such as case-based reasoning approaches \cite{kim_cbm}, where it is necessary to identify clear prototypes to represent each case.

\subsubsection*{Acknowledgments:}
This research is funded by the the University of Stirling CONTEXT research programme and by Bambu (B2B Robo advisor, Singapore). 
%
%
%
%

\end{document}